\documentclass[letterpaper, 10 pt, journal, twoside]{ieeetran} 
\usepackage{times}
\usepackage{epsfig}
\usepackage{graphicx}
\usepackage{amsmath}
\usepackage{amsfonts}
\usepackage{amssymb}
\usepackage{bm}
\usepackage{comment}
\usepackage{longtable}
\usepackage{hhline}
\usepackage{ctable} 
\usepackage{multirow}
\usepackage{caption}
\usepackage{cite}

\usepackage{xcolor}

\DeclareMathOperator*{\argmin}{argmin} 
\DeclareMathOperator*{\argmax}{argmax} 

\usepackage[pagebackref=true,breaklinks=true,letterpaper=true,colorlinks,bookmarks=false]{hyperref}

\IEEEoverridecommandlockouts                              

\title{
Probabilistic Crowd GAN: Multimodal Pedestrian Trajectory Prediction using a Graph Vehicle-Pedestrian Attention Network
}

\author{Stuart Eiffert, Kunming Li, Mao Shan, Stewart Worrall, Salah Sukkarieh and Eduardo Nebot

\thanks{Manuscript received: Feb, 24, 2020; Revised May, 11, 2020; Accepted June, 6, 2020. This paper was recommended for publication by Editor Youngjin Choi upon evaluation of the Associate Editor and Reviewers' comments.}
\thanks{This  work  has  been  funded  by  the  Australian  Centre  for Field Robotics (ACFR), University of Michigan / Ford Motors Company Contract ``Next generation Vehicles”,  Transport for New South Wales (TfNSW) and iMOVE CRC and supported by the  Cooperative  Research  Centres  program, an Australian Government initiative.} \thanks{All authors are with ACFR, Sydney University}
\thanks{Stuart Eiffert and Kunming Li are co-first authors}
\thanks{(\tt\footnotesize s.eiffert@acfr.usyd.edu.au; \tt\footnotesize k.li@acfr.usyd.edu.au)}%
        
\thanks{Digital Object Identifier (DOI): see top of this page.}

}

\begin{document}

\setlength{\abovedisplayskip}{8pt}
\setlength{\belowdisplayskip}{6pt}



\IEEEpubid{\begin{minipage}[t]{\textwidth}\ \\[6pt]
        \centering\normalsize{Copyright \copyright 2020 IEEE. Personal use is permitted. For any other purposes, permission must be obtained from the IEEE by emailing pubs-permissions@ieee.org.}
\end{minipage}}         


\markboth{IEEE Robotics and Automation Letters. Preprint Version. Accepted June, 2020 \url{http://dx.doi.org/10.1109/LRA.2020.3004324}}
{Eiffert \MakeLowercase{\textit{et al.}}: Probabilistic Crowd GAN}  

\maketitle

\begin{abstract}

Understanding and predicting the intention of pedestrians is essential to enable autonomous vehicles and mobile robots to navigate crowds. This problem becomes increasingly complex when we consider the uncertainty and multimodality of pedestrian motion, as well as the implicit interactions between members of a crowd, including any response to a vehicle. Our approach, Probabilistic Crowd GAN, extends recent work in trajectory prediction, combining  Recurrent Neural Networks (RNNs) with Mixture Density Networks (MDNs)  to output probabilistic multimodal predictions, from which likely modal paths are found and used for adversarial training. We also propose the use of Graph Vehicle-Pedestrian Attention Network (GVAT), which models social interactions and allows input of a shared vehicle feature, showing that inclusion of this module leads to improved trajectory prediction both with and without the presence of a vehicle. Through evaluation on various datasets, we demonstrate improvements on the existing state of the art methods for trajectory prediction and illustrate how the true multimodal and uncertain nature of crowd interactions can be directly modelled.

\end{abstract}

\section{Introduction}
\begin{figure}
    \centering
	\includegraphics[width=8.2cm,height=8.5cm]{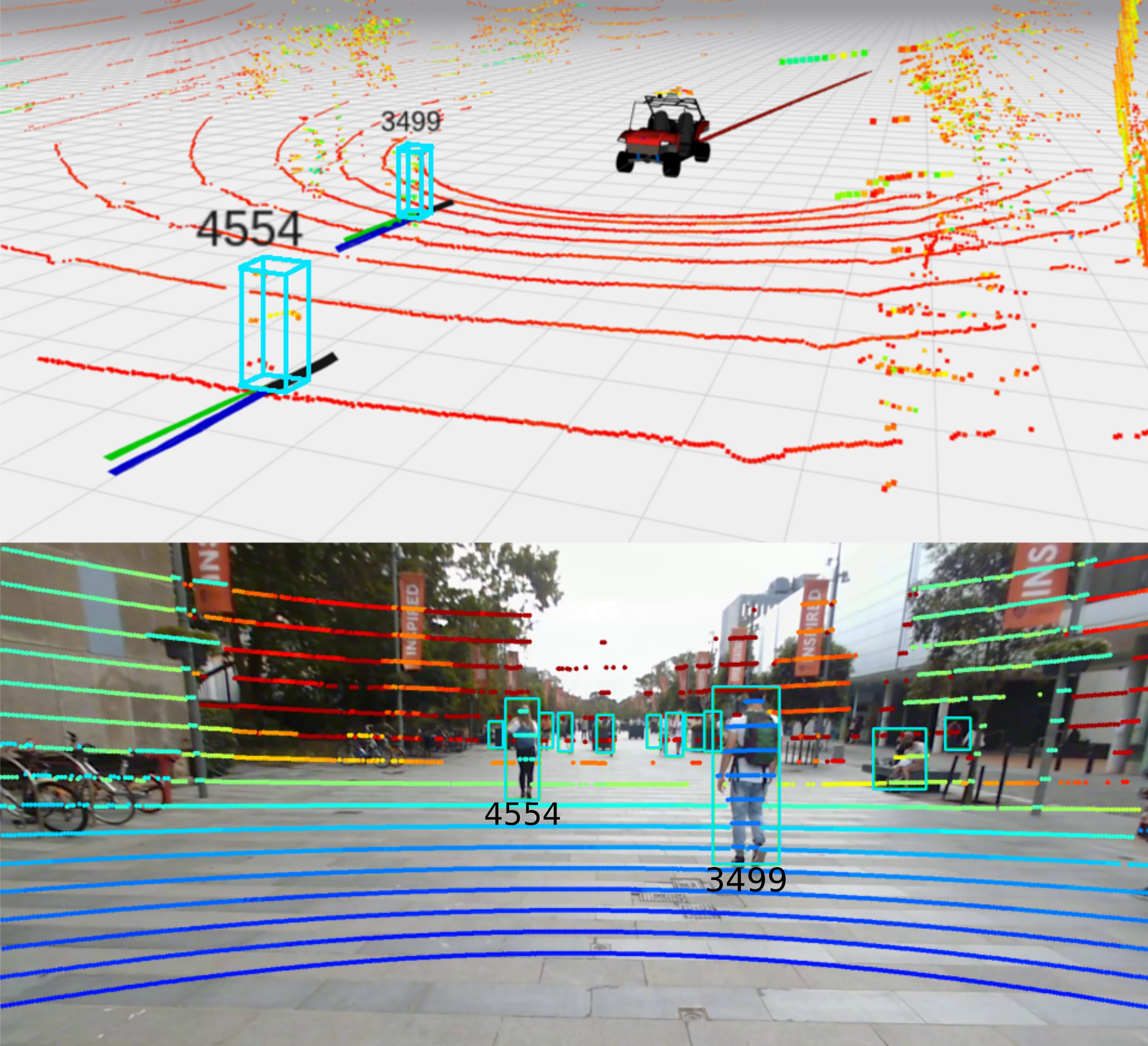}
	\caption{\textit{Motion of detected pedestrians is predicted using our method Probabilistic Crowd GAN with a Graph Vehicle-Pedestrian Attention Network (PCGAN). Observed trajectories are shown in black. The most likely modal path of the multimodal probabilistic prediction are shown in green against ground truth in blue. Predictions use the vehicle's motion as a feature input on the USyd Campus dataset \cite{usydc}.}}
	\label{front_image}
\end{figure}

\IEEEPARstart{P}{edestrian} motion prediction is required for the safe and efficient operation of autonomous vehicles and mobile robots in shared pedestrian environments, such as malls and campuses, as shown in \textit{Fig. \ref{front_image}}. 
The motion of individual members of a crowd is dependent on the motion of others nearby, including any vehicles, and contains significant uncertainty during interactions. 
In order to better predict pedestrian motion we need to be able to model this uncertainty, which is often multimodal due to the variety of ways in which individuals can interact and avoid each other.
Recent works that aim to capture this multimodal and probabilistic nature of crowd interactions have attempted to do so through repeated sampling of generative models, often using Recurrent Neural Networks (RNNs)-based autoencoders trained as Generative Adversarial Networks (GANs) \cite{gupta2018social, kosaraju2019social}.
Due to the nature of adversarial training, where generated trajectories must match the form of the ground truth for comparison by the Discriminator, these methods are limited to generating non-probabilistic outputs. Instead, they require repeated sampling with use of a random latent variable to identify the true multimodal distribution during inference. 

Additionally, in applications involving the use of a single-vehicle around pedestrians, such as autonomously navigating a university campus, accurate prediction of nearby pedestrian motion requires inclusion of vehicle-pedestrian interactions in any predictive model. Recent work \cite{kosaraju2019social} has shown that the use of Graph Attention Networks (GATs) \cite{velivckovic2017graph} can improve the modelling of social interactions between pedestrians, as compared to previously used social pooling layers.

\newpage
Our proposed method, Probabilistic Crowd GAN (PCGAN), allows for the direct prediction of probabilistic multimodal outputs during adversarial training.  We make use of a Mixture Density Network (MDN) within the GAN's generator to output a Gaussian mixture model (GMM) for each pedestrian, demonstrating how clustering of each component of the GMM allows the finding of likely modal-paths, that can then be compared to ground truth trajectories by the GAN's discriminator. Additionally, we extend the use of GATs for modelling crowd interactions to include heterogeneous interactions between a vehicle and pedestrians, as a Graph Vehicle-Pedestrian Attention Network (GVAT) used for modelling social interactions in our method. We validate our approach on several publicly available real world datasets of pedestrian crowds, as well as two datasets which include crowd-vehicle interactions.

Our main contributions in this work include:
\begin{itemize}
    \item Direct multimodal probabilistic output from a GAN for trajectory prediction.
    \item Extension of Graph Attention Networks to include a shared vehicle feature in the pooling mechanism.
    \item Improved pedestrian motion prediction both with and without the presence of a single vehicle.
\end{itemize}

\section{Related Work}
\textbf{Pedestrian Trajectory Prediction:}
Approaches to motion prediction in crowds have tended to focus either on modelling scene-specific motion patterns through the inclusion of contextual features \cite{lee2017desire} and learning crowd motion for a specific observed scene \cite{yi2015understanding,zhi2019kernel}, or on interactions between individuals. Crowd interactions have been modelled using either hand-crafted methods such as the Social Force Model (SFM) \cite{socialforce}, or using learnt models of interaction. 
Recent developments in learning-based methods of trajectory prediction such as RNNs \cite{alahi2016social,Vemula2018} allow for improved prediction in crowded environments, outperforming parametric based methods such as SFM \cite{Becker2018}. These methods have been applied to multimodal prediction by learning semantically meaningful latent representations in conditional variational autoencoders \cite{Salzmann2020, Hu2019} and GANs \cite{Huang2019}, or through clustering modal paths in output distributions \cite{zyner2019naturalistic}. However, these methods can still fail to outperform even simple baselines such as constant velocity models in many situations \cite{Scholler2019}.

\textbf{GANs for Probabilistic Prediction:}
GANs \cite{NIPS2014_5423} have been recently used to enable the generation of socially acceptable trajectories in crowd motion prediction. Gupta et al. \cite{gupta2018social} proposed Social GAN, in which the generator of the network consists of an LSTM based encoder-decoder with a social pooling layer modelling the relationship between each pedestrian. The output trajectories of the generator are directly compared to the ground truth by the Discriminator. 
Social-BiGAT \cite{kosaraju2019social} extends this idea, introducing a flexible graph attention network and further encouraging generalization towards a multi-modal distribution. This method, as well as a similar GAN based approach proposed by \cite{sophie2018}, also make use of overhead contextual scene inputs, which are often difficult to capture in autonomous driving systems.  
Prior work \cite{gupta2018social,lee2017desire,Huang2019} using GANs for trajectory prediction has followed the assumption from GAN application to image synthesis that we cannot efficiently evaluate the output distribution, but can sample from it, requiring multiple iterations to identify the true multimodal distribution. However, our problem's output distribution is much lower-dimensional than image synthesis, and has been modelled previously by GMMs as in \cite{ivanovic2019, Salzmann2020, zyner2019naturalistic}, allowing a distribution to be generated from a single iteration. Further, our aim differs from the synthesis in that we are not trying to just generate samples in the style of ground truth conditioned on an observation, but rather samples that mimic ground truth.

\textbf{Interaction Modelling:}
Alahi et al. \cite{alahi2016social} proposed the use of RNNs with a social pooling layer to capture interactions between pedestrians in a crowd, with  similar pooling layer being used in Social GAN \cite{gupta2018social}. The pooling mechanism used in \cite{Chandra_2019_CVPR} allows interactions between different agent types by learning respective weightings for each relationship. Recent works \cite{Eiffert2019,ma2019trafficpredict} have extended the work of Vemula et al.\cite{Vemula2018} to apply Structural-RNNs \cite{Jain_2016_srnn} to heterogeneous interactions, modelling multiple road agents types using RNNs in a spatio-temporal graph.
 Veli{\v{c}}kovi{\'c} proposed graph attention networks (GAT) \cite{velivckovic2017graph} to implicitly assign different importance to nodes in graph structured data. Kosaraju et al. \cite{kosaraju2019social} applied this concept to multimodal trajectory prediction by formulating pedestrian interactions as a graph, however apply the graph structure only within the pooling mechanism as a GAT, rather than modelling each relationship of the graph as a separate RNN as in \cite{Vemula2018}. We extend this idea in our work to demonstrate how a vehicle's feature can be included in GAT as a GVAT.

\section{Method}
\subsection{Problem Definition}
In this paper, we address the problem of pedestrian trajectory prediction in crowds both with and without the presence of a vehicle. Given observed trajectories $\textbf{X}$, and the vehicle path $\textbf{V}$, for all time steps in period $ t \leq T_{obs}$, where $\textbf{X}^t= [\textbf{X}_1^t,\textbf{X}_2^t...,\textbf{X}_N^t]$ for $N$ pedestrians within a scene, our aim is to predict the likely future trajectories $ \bar{\textbf{Y}}^t= [\bar{\textbf{Y}}_1^t,\bar{\textbf{Y}}_2^t...,\bar{\textbf{Y}}_N^t]$ for each pedestrian in $N$, across a future time period $T_{obs} < t \leq T_{pred}$. The input position of the $i$th pedestrian at time $t$ is defined as $\textbf{X}_{i}^t = (x_{i}^{t},y_{i}^t)$ and the vehicle as $\textbf{V}^t = (x_{v}^{t},y_{v}^{t})$. 
We denote $\textbf{Y}$ as the ground truth future trajectory, with the position of the $i$th pedestrian at time $t$ defined as $\textbf{Y}_{i}^t = (x_{i}^{t},y_{i}^t)$ and predicted position as $\bar{\textbf{Y}}_{i}^t = \{(\bar{x}_{i}^{t},\bar{y}_{i}^t, \bar{w}_i)_{m=1}^M \} $  for all predicted modal paths $m \in M$, , where $\bar{w}_i^m$ is the likelihood of the predicted modal path $m$ for agent $i$. $\bar{\textbf{Y}}_{i}^t$ is found from the probabilistic output $\hat{\textbf{Y}}_i^t$, a Gaussian mixture model (GMM) detailed in \textit{Eq. \ref{eq:mnd}}.
\begin{figure*}[tbp] 
\centering
\includegraphics[width=17cm,height=4.5cm]{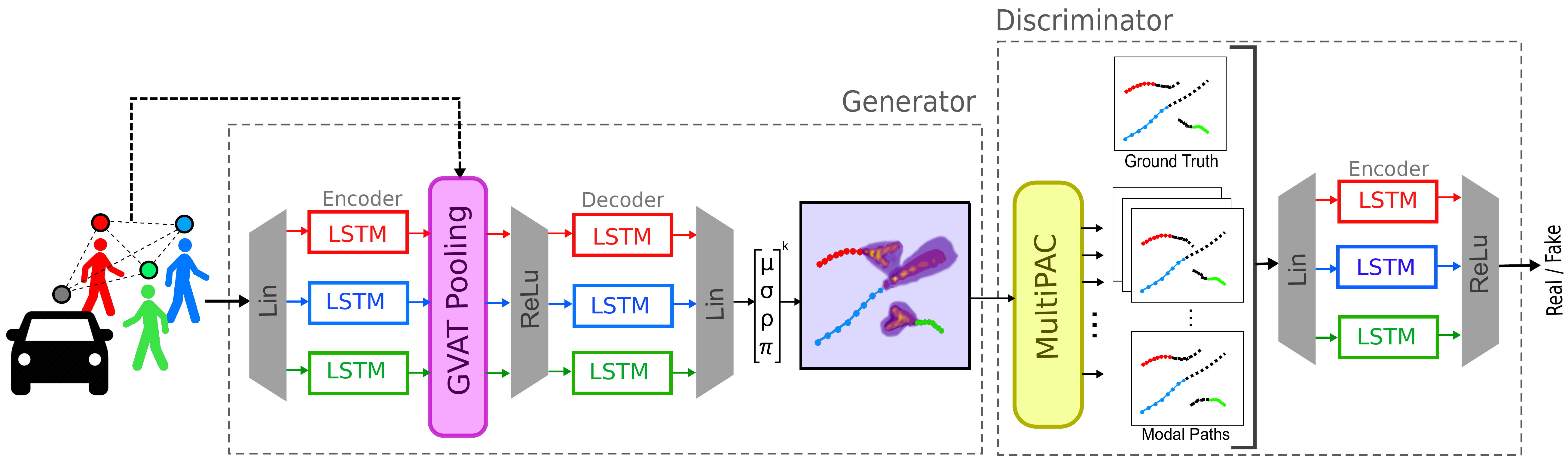}
\caption{\textit{Observed pedestrian trajectories are passed to the Generator's encoder LSTM, whilst the relative position of all agents, including any vehicle, are passed to the GVAT Pooling module. The Generator outputs a GMM for each agent, from which the MultiPAC module finds the likely modal paths, which are compared to ground truth paths by the Discriminator.}}
\label{fig_archi}
\end{figure*}
\subsection{Overview}
Our approach consists of two networks, a Generator and a Discriminator trained adversarially. The Generator is composed of an RNN encoder, our GVAT module, an RNN decoder, and a Mixture Density Network (MDN). The Discriminator is composed of the modal path clustering module MultiPAC, an RNN encoder and a multilayer perceptron (MLP).  
\textit{Fig.\ref{fig_archi}} illustrates the overall system architecture.

\textbf{Generator:}
The Generator is based on an RNN encoder-decoder framework using LSTM modules, where the GVAT Pooling module is applied to the hidden states between the encoder and decoder LSTMs. The input to the encoder LSTM at each timestep $ t < T_{obs}$ is the observed position of each pedestrian $i\in N$, which is first passed through a linear embedding layer $\phi_{e}$ as follows:
\begin{align}
e^{t}_{i} = \phi_{e}(x^{t}_{i},y^{t}_{i}; W_{emb}^e)\\   
h^{t}_{ei} = LSTM(h^{t-1}_{ei}, e^{t}_{i};W_{enc})\label{eq:encoder} 
\end{align}

where $W_{emb}$ is embedding weight of $\phi_{e}$. All pedestrians within a scene share the LSTM weights $W_{enc}$. 
The decoder's initial hidden state at $t=T_{obs}$ is composed of the encoder's final hidden state, concatenated with the transformed output of GVAT Pooling for each agent, detailed further in \textit{Section \ref{sec_gvat}}. The first input to the decoder at $t=T_{obs}$ is again the observed pedestrian positions, passed first through a linear embedding layer $\phi_{d}$ in the same form as $\phi_{e}$ with separate weights. However, as the decoder outputs are a distribution, rather than a single point, we do not simply pass the prediction from the prior timestep as input to the decoder's current timestep. Instead, for all prediction timesteps $T_{obs} < t \leq T_{pred}$ the decoder inputs are zeros. This is done as opposed to other probabilistic approaches which feed a sample from the prior output as current input to the decoder.
This \textit{zero-feed} approach is performed for both training and inference, and has been shown to improve performance for probabilistic outputs \cite{zyner2019naturalistic}.

\begin{align}
g^{t}_{i} &= GVAT(\textbf{X}^{t} ,\textbf{V}^{t}, h_{e}^t)\\ \label{eq:GVAT}
h^{t}_{di} &= LSTM(MLP_{dec}(h^{t-1}_{di},g^t_{i}),d^{t}_{i};W_{dec})\\  \label{eq:decoder}
&{d^{t}_{i}} =
\begin{cases}
\phi_{d}(x^{t}_{i},y^{t}_{i}; W_{emb}^d), & t \leq T_{obs} \nonumber\\
0,& t>T_{obs}
\end{cases}
\end{align}

where $h_{di}^{T_{obs}} = h_{ei}^{T_{obs}}$ and $h_e^t$ is  the combined output of Eq.\ref{eq:encoder} for all agents in the scene. $MLP_{dec}$ is a multi-layer perception with ReLU non-linearity and $W_{dec}$ is the embedding weight.
The outputs of the decoder are passed through a linear embedding layer $\phi_{mdn}$ with weights $W_{mdn}$ that maps to  a bivariate GMM output $\hat{Y}^t_i$ for each agent's position at each predicted timestep. $\hat{Y}^t_i$ is then passed to the MultiPAC module to determine the set of likely modal paths $\bar{Y}^t_i$:
\begin{align}
    \hat{Y}^t_i= \phi_{mdn}(h^{t}_{di}; W_{mdn}) 
\end{align}

\textbf{Discriminator:}
The Discriminator is comprised of a MultiPAC module, and an LSTM encoder of the same form as the Generator's, with separate weights. The output of Generator $\hat{Y}^t_i$ is first passed to MultiPAC, from which we compute the set of likely modal paths $\bar{Y}^t_i$, as detailed in Section \ref{section:mpc}. This produces trajectories in the same form as the ground truth, allowing comparison by the Discriminator's encoder. The encoder is applied across all timesteps $0 < t \leq T_{pred}$, with inputs first passed through a linear embedding layer.
Outputs of the encoder are passed to a multilayer perceptron (MLP) with ReLU activation, classifying the path as either a Real or Fake.

\textbf{Loss:}
Training of the network is achieved using two loss functions $L_{lh}$ and $L_{adv}$.
$L_{lh}$ is the negative log-likelihood of the ground truth path $\textbf{Y}$  given the Generator $G$'s output $\hat {\textbf{Y}}$, across all prediction timesteps, for all pedestrians:
\begin{equation}
L_{lh}=-\sum_{t=T_{obs}+1}^{T_{pred}}\sum_{i}^{N} \log (P({Y_{i}^{t}}|\hat{Y_{i}^{t}}))
\end{equation}
$L_{adv}$ is the adversarial loss, determined from the binary cross entropy of the Discriminator $D$'s classification of the modal paths $\bar {\textbf{Y}}$ produced from $\hat {\textbf{Y}}$ by MultiPAC:
\begin{equation}
L_{adv}=\mathbb{E}[log(D(X_i,Y_i))] + \sum_{m=1}^{M}\mathbb{E}[log(1-w^m_iD(X_i,\bar{Y_i})] \label{eq:Ladv}
\end{equation}
where the first term refers to $D$'s estimate of the probability that the ground truth trajectory $Y_i$ is real, and the second term is the sum of weighted estimates for each modal path in the set $\bar{Y}_i $ being real.
We combine the losses to find the optimal Discriminator $D*$ and Generator $G*$, with weighting $\alpha$ applied to $L_{lh}$:
\begin{equation}
G*,D* = \argmin_G \argmax_D [L_{adv} + \alpha L_{lh} ] \label{eq:argminmax}
\end{equation}
\subsection{Graph Vehicle-Pedestrian Attention Network} \label{sec_gvat}
\begin{figure}[t]
\begin{center}
    \includegraphics[width=7.5cm,height=4.0cm]{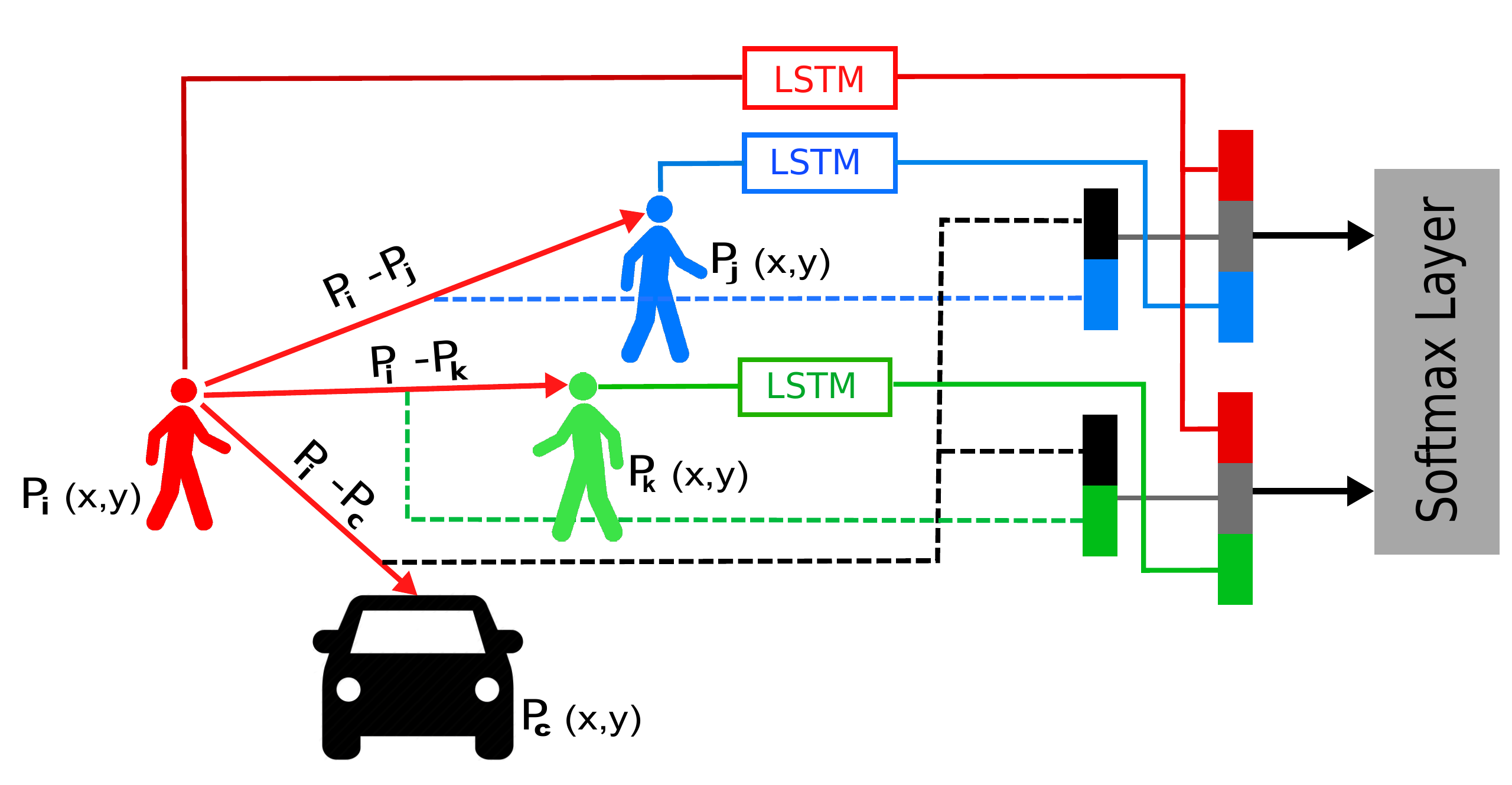}
\end{center}
   \caption{\textit{Node features of agent \textit{i} (red) in GVAT.  The distance from \textit{i} to the vehicle is appended to each other ped-ped distance input before encoding to account for the impact of the vehicle on \textit{i}'s relationships within the graph. The input to softmax layer is $u_{i,j}^{t}$ as per Eq. \ref{eq:att_embed} }}
\label{fig:gvatinputs}
\end{figure}

We introduce a novel Graph Vehicle-Pedestrian Attention Network (GVAT), which extends upon the use of GATs \cite{velivckovic2017graph} for trajectory prediction by \cite{kosaraju2019social} \cite{zhang2019sr}, allowing the modelling of social interactions between all pedestrians in a scene, and accommodating the inclusion of a vehicle if present.
As opposed to \cite{kosaraju2019social}, where only agent hidden states form the GAT input features, we also utilise distance between agents, so that vehicle distance to agent \textit{i} can be included to allow the attention module to account for the impact that the vehicle's motion has on each ped-ped relationship. \textit{Fig. \ref{fig:gvatinputs}}. details the input features of a single node in the graph. 

For the $i$th pedestrian, the input to the softmax layer is formulated across all other pedestrians \newline {j $\in N \setminus \{ i\} $} by embedding the distance from pedestrian $i$ to the neighbour pedestrian $j$ and the vehicle. The softmax scalar $a_{i,j}^{t}$ is then used to scale the amount agent j's hidden state influences agent i. The summed output across all other agents, $g_{i}^{t}$, is then concatenated with i's original state to form the output of GVAT pooling $h_{gi}^{t}$. $\phi_{r}$, $\phi_{u}$ and $\phi_{gat}$ are linear embedding functions, $W_{r}$, $W_{u}$ and $W_{gat}$ denote their parameters respectively:
\begin{align}
r_{i,j}^{t}&=\phi_{r}(x_{i}^{t}-x_{j}^{t},y_{i}^{t}-y_{j}^{t},z^t_i; W_{r})\label{eq:nodeinputs}\\
&{z^{t}_{i}} =
\begin{cases}
(x_{i}^{t}-x_{v}^{t},y_{i}^{t}-y_{v}^{t}), & \textrm{\textit{vehicle present}} \nonumber\\
(0,0),& \textrm{\textit{no vehicle}} \\
\end{cases} \\
u_{i,j}^{t} &= \phi_{u}(concat(r_{i,j}^{t};h_{ej}^{t};h_{ei}^{t});W_{u})\label{eq:att_embed}  \\
a_{i,j}^{t} &= \frac{exp(u_{i,j}^{t})}{\sum\limits_{k \in N \setminus \{ i\}} exp(u_{i,k}^{t})} \label{eq:att_product}\\
g_{i}^{t} &= \sum\limits_{j \in N \setminus \{ i\} } \phi_{gat}(a_{i,j}^t  \cdot h_{ej}^{t};W_{gat}) \label{eq:att_fin} \\
h_{gi}^{t} &= concat(h_{i}^{t}, g_{i}^{t})
\end{align}

\begin{table*}[] \label{table_ex1}
\centering 
\begin{tabular}{c|c|c|c|c|c|c|c|}
\cline{1-8}
\multicolumn{1}{|l|}{\multirow{2}{*}{\textbf{Metric}}}     & \multirow{2}{*}{\textbf{Dataset}} & \multirow{2}{*}{\textbf{Lin}} & \multirow{2}{*}{\textbf{CVM}} & \multirow{2}{*}{\textbf{SGAN}} & \multirow{2}{*}{\textbf{SRLSTM}} & \multicolumn{2}{c|}{Ours}                   \\
\multicolumn{1}{|l|}{}                               &                                   &                                  &                               &                                &                                  & \textbf{PSGAN}       & \textbf{PCGAN}      \\ \hline \hline
\multicolumn{1}{|c|}{\multirow{5}{*}{\textbf{ADE}}} & \textbf{ETH-Univ}                 & 0.50 / 0.79                      & 0.48 / 0.70                    & 0.51 / 0.81                    & \textbf{0.43} / \textbf{0.65}                         & 0.45 / 0.68          & \textbf{0.43} / \textbf{0.65} \\
\multicolumn{1}{|c|}{}                              & \textbf{ETH-Hotel}                & 0.35 / 0.39                      & 0.28 / \textbf{0.33}          & 0.55 / 0.67                    & \textbf{0.24} / 0.42                         & 0.52 / 0.64          & 0.59 / 0.64          \\
\multicolumn{1}{|c|}{}                              & \textbf{UCY-Univ}                 & 0.56 / 0.82                      & \textbf{0.34} / 0.56          & 0.56 / 0.78                    & 0.38 / \textbf{0.53 }                        & \textbf{0.34} / 0.55 & 0.49 / 0.57          \\
\multicolumn{1}{|c|}{}                              & \textbf{UCY-Zara01}               & 0.41 / 0.62                      & 0.28 / 0.46                   & 0.46 / 0.63                    & 0.28 / 0.43                         & \textbf{0.25} / 0.43          & \textbf{0.25 / 0.40}  \\
\multicolumn{1}{|c|}{}                              & \textbf{UCY-Zara02}               & 0.53 / 0.77                      & 0.23 / 0.35          & 0.35 / 0.56                    & 0.24 / \textbf{0.32}                & 0.27 / 0.37          & \textbf{0.22} / 0.34 \\ \hline
\multicolumn{1}{|c|}{\multirow{5}{*}{\textbf{FDE}}} & \textbf{ETH-Univ}                 & 0.88 / 1.57                      & 0.87 / 1.34                   & 0.95 / 1.72                    & \textbf{0.80} / 1.26                & 0.84 / 1.34          & 0.81 / \textbf{1.25} \\
\multicolumn{1}{|c|}{}                              & \textbf{ETH-Hotel}                & 0.60 / 0.72                      & \textbf{0.40 / 0.62}           & 0.49 / 1.71                    & 0.45 / 0.90                         & 1.13 / 1.45          & 1.10 / 1.40           \\
\multicolumn{1}{|c|}{}                              & \textbf{UCY-Univ}                 & 1.01 / 1.59                      & \textbf{0.71} / 1.20           & 1.20 / 1.70                   & 0.81 / \textbf{1.17}                 & \textbf{0.71} / 1.23          & 0.89 / 1.24          \\
\multicolumn{1}{|c|}{}                              & \textbf{UCY-Zara01}               & 0.74 / 1.21                      & 0.57 / 0.99                   & 0.99 / 1.38                    & 0.60 / 0.93                         & 0.53 / \textbf{0.87}          & \textbf{0.53} / 0.89 \\
\multicolumn{1}{|c|}{}                              & \textbf{UCY-Zara02}               & 0.95 / 1.48                      & 0.47 / 0.75          & 0.75 / 1.21                    & 0.51 / \textbf{0.73}                         & 0.56 / 0.76          & \textbf{0.45} / 0.77 \\ \hline
\multicolumn{1}{|c|}{\multirow{5}{*}{\textbf{MHD}}} & \textbf{ETH-Univ}                 & 0.48 / 0.66                      & 0.40 / 0.57                    & 0.44 / 0.66                    & \textbf{0.38 / 0.54}                    & 0.40 / 0.59          & \textbf{0.38} / 0.55 \\
\multicolumn{1}{|c|}{}                              & \textbf{ETH-Hotel}                & 0.33 / 0.33                      & \textbf{0.20 / 0.27}           & 0.22 / 0.69                    & 0.22 / 0.37                       & 0.45 / 0.56          & 0.51 / 0.55          \\
\multicolumn{1}{|c|}{}                              & \textbf{UCY-Univ}                 & 0.52 / 0.76                      & 0.31 / 0.48          & 0.48 / 0.67                    & 0.34 / \textbf{0.45}                               & \textbf{0.30} / 0.49           & 0.41 / 0.50           \\
\multicolumn{1}{|c|}{}                              & \textbf{UCY-Zara01}               & 0.39 / 0.55                      & 0.24 / 0.40                    & 0.40 / 0.52                     & 0.26 / 0.36                             & \textbf{0.23} / 0.37          & \textbf{0.23 / 0.35} \\
\multicolumn{1}{|c|}{}                              & \textbf{UCY-Zara02}               & 0.47 / 0.71                      & 0.23 / \textbf{0.31}                   & 0.31 / 0.49                    & 0.22 / \textbf{0.31}                               & 0.25 / 0.33          & \textbf{0.20 / 0.31} \\ \hline
\end{tabular}
\caption{\textit{Quantitative results of tested methods on all non-vehicle datasets. For each dataset, we compare results across two prediction lengths of 8 and 12 timesteps (3.2 and 4.8 secs), showing Average Displacement Error (ADE), Final Displacement Error (FDE), and Modified Hausdorff Distance (MHD) in meters.}\strut}
	\label{tab:1}
\end{table*}

\begin{table}[]
\setlength\tabcolsep{2.6pt} 
\begin{tabular}{cccccccc}
\hline
\multicolumn{1}{|c|}{\multirow{2}{*}{\textbf{Metrics}}} & \multicolumn{1}{c|}{\multirow{2}{*}{\textbf{Dataset}}} & \multicolumn{1}{c|}{\multirow{2}{*}{\textbf{Lin}}} & \multicolumn{1}{c|}{\multirow{2}{*}{\textbf{CVM}}} & \multicolumn{1}{c|}{\multirow{2}{*}{\textbf{SGAN}}} & \multicolumn{1}{c|}{\multirow{2}{*}{\textbf{SRLSTM}}} & \multicolumn{2}{c|}{Ours}                                                          \\
\multicolumn{1}{|c|}{}                                  & \multicolumn{1}{c|}{}                                  & \multicolumn{1}{c|}{}                              & \multicolumn{1}{c|}{}                              & \multicolumn{1}{c|}{}                               & \multicolumn{1}{c|}{}                                 & \multicolumn{1}{c|}{\textbf{PSGAN}} & \multicolumn{1}{c|}{\textbf{PCGAN}} \\ \hline \hline
\multicolumn{1}{|c|}{\multirow{2}{*}{\textbf{ADE}}}     & \multicolumn{1}{c|}{\textbf{USyd}}                     & \multicolumn{1}{c|}{0.16}                          & \multicolumn{1}{c|}{0.13}                          & \multicolumn{1}{c|}{0.16}                           & \multicolumn{1}{c|}{\textbf{0.11}}                    & \multicolumn{1}{c|}{\textbf{0.11}}           & \multicolumn{1}{c|}{\textbf{0.11}}  \\
\multicolumn{1}{|c|}{}                                  & \multicolumn{1}{c|}{\textbf{VCI}}                      & \multicolumn{1}{c|}{0.11}                          & \multicolumn{1}{c|}{0.09}                          & \multicolumn{1}{c|}{0.12}                           & \multicolumn{1}{c|}{\textbf{0.08}}                    & \multicolumn{1}{c|}{0.12}                    & \multicolumn{1}{c|}{\textbf{0.08}}  \\ \hline
\multicolumn{1}{|c|}{\multirow{2}{*}{\textbf{FDE}}}     & \multicolumn{1}{c|}{\textbf{USyd}}                     & \multicolumn{1}{c|}{0.30}                          & \multicolumn{1}{c|}{0.24}                          & \multicolumn{1}{c|}{0.31}                           & \multicolumn{1}{c|}{0.22}                             & \multicolumn{1}{c|}{\textbf{0.21}}           & \multicolumn{1}{c|}{\textbf{0.21}}  \\
\multicolumn{1}{|c|}{}                                  & \multicolumn{1}{c|}{\textbf{VCI}}                      & \multicolumn{1}{c|}{0.23}                          & \multicolumn{1}{c|}{0.18}                          & \multicolumn{1}{c|}{0.22}                           & \multicolumn{1}{c|}{0.16}                             & \multicolumn{1}{c|}{0.20}                    & \multicolumn{1}{c|}{\textbf{0.15}}  \\ \hline
\multicolumn{1}{|c|}{\multirow{2}{*}{\textbf{MHD}}}     & \multicolumn{1}{c|}{\textbf{USyd}}                     & \multicolumn{1}{c|}{0.12}                          & \multicolumn{1}{c|}{0.09}                          & \multicolumn{1}{c|}{00.12}                          & \multicolumn{1}{c|}{0.09}                             & \multicolumn{1}{c|}{\textbf{0.08}}           & \multicolumn{1}{c|}{0.09}           \\
\multicolumn{1}{|c|}{}                                  & \multicolumn{1}{c|}{\textbf{VCI}}                      & \multicolumn{1}{c|}{0.09}                          & \multicolumn{1}{c|}{0.07}                          & \multicolumn{1}{c|}{0.09}                           & \multicolumn{1}{c|}{\textbf{0.07}}                    & \multicolumn{1}{c|}{0.09}                    & \multicolumn{1}{c|}{\textbf{0.07}}  \\ \hline
\end{tabular}
\caption{\textit{Quantitative results of tested methods on both vehicle datasets. We compare results using a prediction length of 12 timesteps (1.0 second (VCI) and 1.2 seconds (USyd)). ADE, FDE and MHD are shown in meters.}\strut}
	\label{tab:2}
\end{table}
\subsection{Mixture Density Network}
An MDN is used to allow the Generator to propose a multimodal solution for each agent's future trajectory, with assigned relative likelihoods for each Gaussian component of the mixture model. To achieve this, the output of the Generator's decoder is passed through a multilayer perceptron (MLP) to produce output $\hat{Y}^t$ in the form: 
\begin{equation}
\hat{Y}^t=[{\pi, \mu_x^k, \mu_y^k, \sigma_x^k, \sigma_y^k, \rho^k}_{k=1}^K]^t \label{eq:mnd}
\end{equation}
where $K$ is the total number of components used in the mixture model, $\pi$ is the weight of each component in the mixture, $\mu$ is the mean and $\sigma$ the standard deviation per dimension, and $\rho$ is the correlation coefficient, for each timestep $T_{obs} < t \leq T_{pred}$. This is performed separately for each agent $i\in N$, which has been left off for clarity.

\subsection{Modal Path Clustering}
\label{section:mpc}
In order to allow the training of the Discriminator, the output of the Generator must be converted to the same form as the ground truth trajectories $\textbf{Y}$.
This requires extracting individual tracks from the GMM $\hat{\textbf{Y}}$, whilst preserving the multimodality of the distribution.
We achieve this by adapting the multiple prediction adaptive clustering algorithm (MultiPAC) proposed by Zyner et al. \cite{zyner2019naturalistic} to allow backpropagation for use during training. 
MultiPAC finds the set of likely `modal paths' $\bar {\textbf{Y}}$, for each pedestrian from $\hat {\textbf{Y}}$. It achieves this by clustering the components of the GMM  at each timestep using DBSCAN \cite{ester1996}, determining each cluster's centroid from the weighted average of all Gaussians in the mixture. Clusters in subsequent timesteps are assigned to parent clusters, forming a tree of possible paths with an upper limit of children at each timestep being the number of mixtures used within the GMM. This tree is computed from a single forward pass of the model, resulting in a forked trajectory when diverging possible paths are predicted for a single agent, passing each branch of the fork separately to the Discriminator. The paths from each leaf to the root are returned as the set of modal paths $\bar {\textbf{Y}}$ for each pedestrian with assigned likelihoods $w$.

\subsection{Implementation}
The LSTM encoder and decoder of the Generator both have a hidden state size of 32, whilst the Discriminator's LSTM encoder hidden state size is 64.
The linear embedding layers applied to all inputs of both encoders, and the first input of the decoder at $t=T_{obs}$, produce a 16-dimensional vector from the input coordinates. The linear embedding layer at the decoder's output produces a vector of $6\times K$, where $K$ is the number of components in the GMM, set as 6 for all experiments. Both MLPs have a hidden layer of size 64 and use ReLU activation. 
The network is trained initially for 10 epochs using only the negative log-likelihood loss $L_{lh}$, before training adversarially using both loss functions for a further 90 epochs. This initial training is implemented in order to encourage the Generator to produce sensible results before comparison to the ground truth by the Discriminator, and also allows training to converge in significantly fewer iterations.
All training is performed using Adam optimiser with a batch size of 32 and initial learning rate of 0.001. The $\alpha$ weighting applied to $L_{lh}$ in \textit{Eq. \ref{eq:argminmax}} is chosen as 0.1.

\section{Experiments}
We conduct two experiments in order to validate our method's effectiveness both with and without a vehicle feature input.
Firstly, we evaluate our model without any vehicle feature input on two publicly available datasets of real world interacting pedestrian crowds , ETH \cite{pellegrini2009you} and UCY\cite{lerner2007crowds}. 
Next, we verify our model using a vehicle feature input on two datasets of interacting pedestrian crowds and vehicles. These include the publicly available dataset, Vehicle-Crowd Interaction DUT dataset (VCI) \cite{yang2019top}, and the USyd Campus Dataset (USyd) \cite{usydc}.


\subsection{Datasets}
\textbf{ETH and UCY} contain 5 crowd scenes: ETH-Univ, ETH-Hotel, UCY-Zara01, UCY-Zara02, and UCY-Hotel. 
Each dataset is converted to world coordinates with an observation frequency of 2.5 Hz, similar to \cite{gupta2018social}. We deal with the ETH-Univ frame rate issue addressed by \cite{zhang2019sr} similarly by treating every 6 frames as 0.4s rather than 10 frames, and retrain all comparative models for this scene.

\textbf{USyd} is collected on a weekly basis by \cite{usydc} from March 2018 over the University of Sydney campus and surroundings. The dataset contains over 52 weeks drives and covers various environmental conditions. Since our research work primarily focuses on predicting socially plausible future trajectories of pedestrians under the influence of one vehicle, we select from the dataset 17 scenarios in an open large area with high pedestrian activity. Pedestrians are detected by fusing YOLO \cite{2016you} classification results and LiDAR point clouds from vehicle onboard sensors, as illustrated in \textit{Fig. \ref{front_image}}. The GMPHD \cite{Vo2006} tracker is used to automatically label the trajectories of pedestrians. To increase the diversity of data available for training models, we apply data augmentation by flipping 2D coordinates randomly. 
Due to limitations regarding the length of time agents are observed in this dataset, we use an observation frequency of 10 Hz, rather than downsampling to be comparable to Experiment 1.

\textbf{VCI} proposed by \cite{yang2019top}, contains two scenes of labelled video from a birds eye view of vehicle-crowd interactions, recorded at 24 Hz. We downsample this dataset to 12 Hz in order to make results comparable with the USyd dataset. We remove sequences which contain more than one vehicle.

\begin{figure*} \label{qual_nonvehicle}
    \centering
	\includegraphics[width=17.5cm,height=8.1cm]{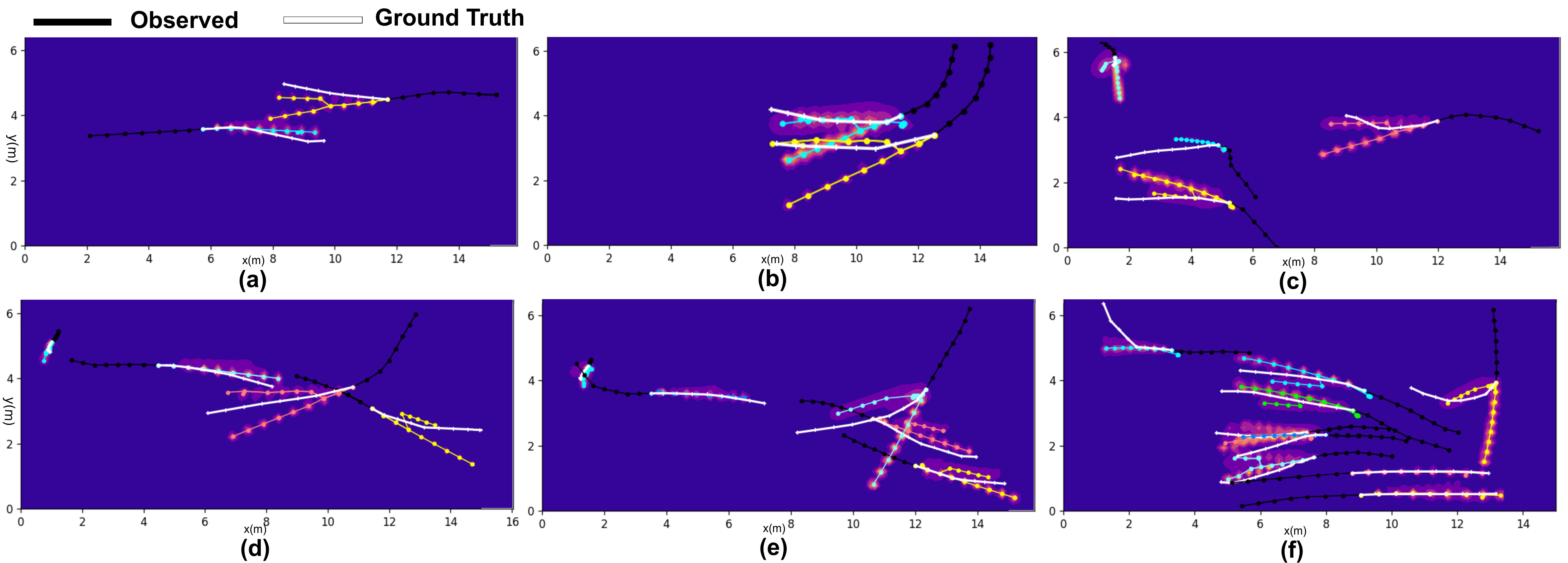}
	\caption{\textit{The predicted modal path trees of MultiPAC are shown in a different colour for each pedestrian, over the probabilistic output of the Generator. Example interactions are from the ETH and UCY datasets, using PCGAN trained without vehicle feature input. Multimodal output is clear in examples in which pedestrians may take one of multiple possible future paths to avoid the collision. Example (a) displays two likely paths that the yellow agent might have taken as the pedestrians approach each other. Example (b) similarly shows multimodal possibilities, including the pedestrians continuing to turn, or to start travelling forwards. Examples (c) through (f) demonstrate similar behaviour in larger pedestrian crowds.}}
	\label{qual_nonvehicle}
\end{figure*}

 \begin{figure*} \label{qual_vehicle}
    \centering
	\includegraphics[width=17.5cm,height=4.5cm]{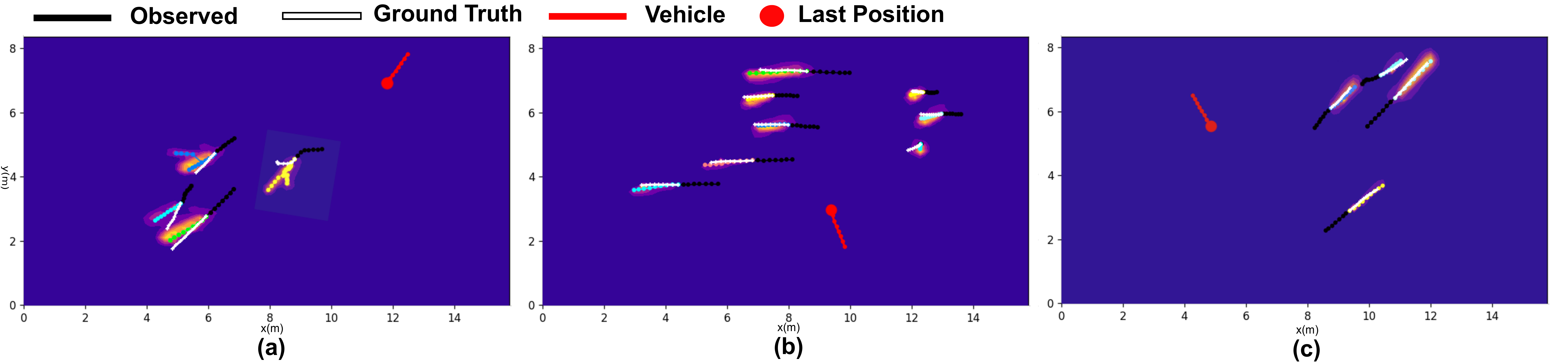}
	\caption{\textit{Predicted pedestrian trajectories using PCGAN trained with vehicle feature input on the VCI dataset. Example (a) illustrates a scene of a vehicle approaching pedestrians from behind, displaying expected multimodal reactions of the pedestrians to either continue forwards at increased speed or move aside. Examples (b) and (c) further illustrate this concept, showing how the direction of the vehicle approach can impact the pedestrians' reaction.}}
	\label{qual_vehicle}
\end{figure*}

\subsection{Evaluation Metrics and Baselines}
\subsubsection{Metrics}
Similar to prior work \cite{alahi2016social} we included two error metrics:  Average Displacement Error (ADE) and Final Displacement Error (FDE). However, as discussed by Zyner et al. \cite{zyner2019naturalistic}, these commonly used measures do not consider outliers throughout the prediction and penalize misalignment in time and space equally. This can result in a prediction with an incorrect speed profile but correct direction having a similar error as a prediction with the completely wrong direction, which is a significantly worse result. As such, Modified Hausdorff Distance (MHD) \cite{odena2017conditional}, which does not suffer this issue, is also included as an evaluation metric.

The metrics used are as follows:
\begin{itemize}
  \item ADE: Average Euclidean distance between ground truth and prediction trajectories over all predicted time steps.
  \item FDE: Euclidean distance between ground truth and prediction trajectories for the final predicted time step.
  \item MHD: A measure of similarity between trajectories, determining the largest distance from each predicted point to any point on the ground truth trajectory.
\end{itemize}

\newpage
\subsubsection{Baseline Comparisons}
We compare our model against the following baseline and state of the art methods:
\begin{itemize}
    \item Lin: A linear regression of pedestrian motion over each dimension. 
    \item CVM: Constant Velocity Model proposed by ~\cite{Scholler2019}.
    \item Social GAN (SGAN) \cite{gupta2018social}: LSTM encoder-decoder with a social pooling layer, trained as a GAN.
    \item SR-LSTM \cite{zhang2019sr}: LSTM based model using a State Refinement module. 
\end{itemize}

Additionally, we perform an ablation study of our method, comparing our model using a social pooling layer as proposed in \cite{gupta2018social} (PSGAN), and the model trained instead using our GVAT module for social pooling (PCGAN). 

As SGAN requires a random noise input for generation, we sample this method 10 times, returning the average error of all samples, as opposed to \cite{gupta2018social}, where the sample with the best error compared to the ground truth was used.

\subsection{Methodology}
For all evaluations on our probabilistic methods, PSGAN and PCGAN, we apply MultiPAC to the output of the Generator to find all modal paths, using the predicted path with the highest probability to compute the error.

\textbf{Experiment 1:}  
Similar to \cite{alahi2016social}, we train on four datasets and evaluate on the remaining one.
We observe the last 8 timesteps of each trajectory (3.2 seconds) and predict for the next 8 (3.2 seconds) and 12 (4.8 seconds) timesteps.\newline

\phantom{xx}\textbf{Experiment 2:}
Each dataset is split into non-overlapping train, validation and test sets in ratios of 60\%, 20\% and 20\% respectively. We observe 8 timesteps of each trajectory (0.67 seconds (VCI) and 0.8 seconds (USyd)) and predict for the next 12 timesteps (1.0 second (VCI) and 1.2 seconds (USyd)).

\begin{figure} \label{qual_results_compare}
    \centering
	\includegraphics[width=8.5cm,height=11cm]{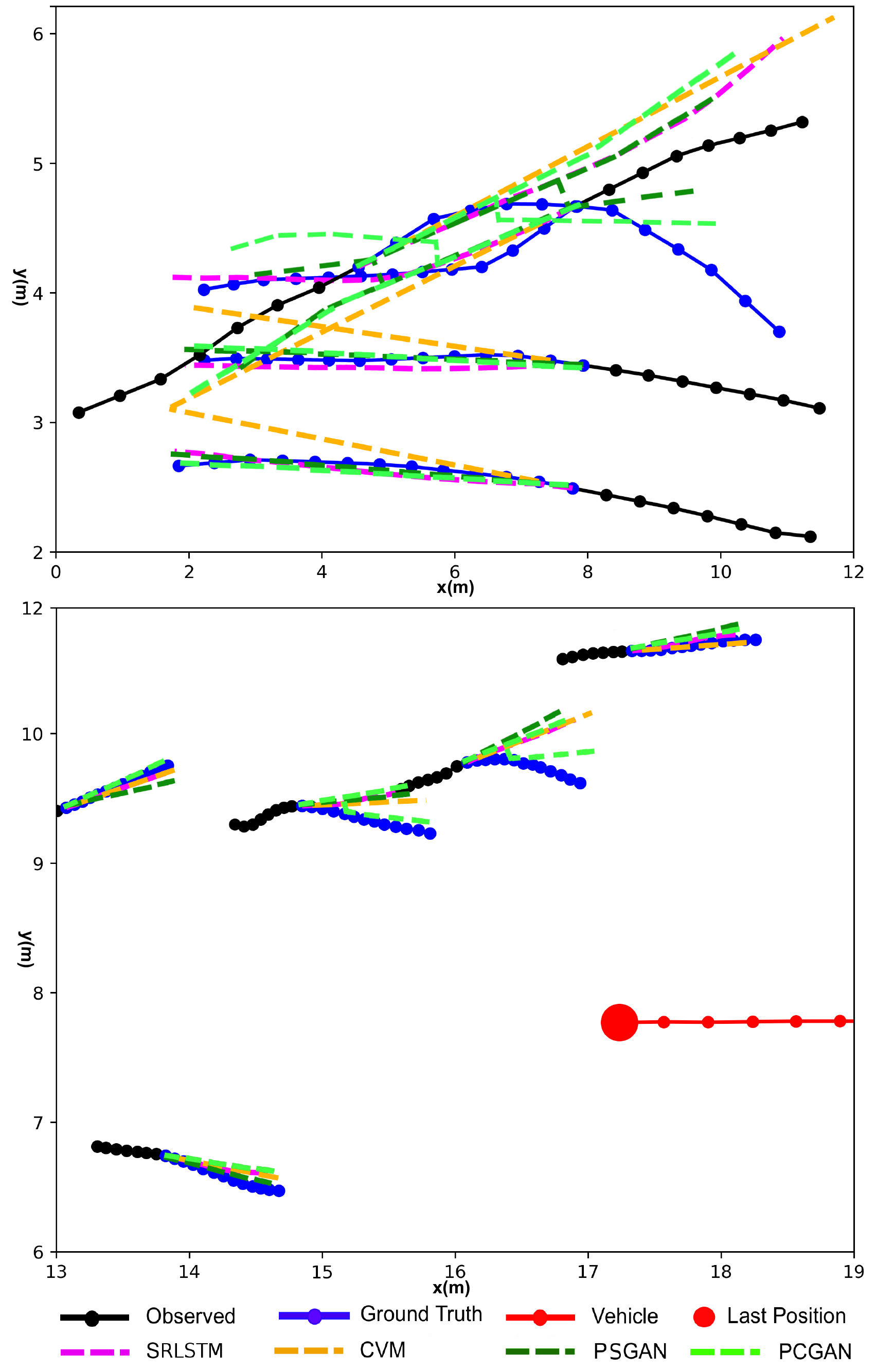}
	\caption{\textit{Comparison of methods on UCY-Zara01 (top) and VCI (bottom) showing the entire modal path tree for both PSGAN and PCGAN. Whilst CVM and SRLSTM outperform our methods on some datasets, our multimodal output better represents uncertainty in crowd interactions, demonstrated in the top example where the possibility that oncoming pedestrians could avoid each other in two different ways is reflected in the branching modal path trees. The bottom illustrates how PCGAN improves predictions in the presence of a vehicle compared to PSGAN, accounting for the impact of the vehicle's motion on pedestrians' motion.}}
	\label{qual_comp}
\end{figure}
\section{Results and Discussion}
\subsection{Quantitative Evaluation}
\textbf{Experiment 1:}  
  \textit{Table.\ref{tab:1}} compares results for all methods on the ETH and UCY datasets. Our adversarial approaches PSGAN and PCGAN clearly outperform the previous sampling-based adversarial approach \cite{gupta2018social} demonstrating that the use of a direct probabilistic generator output can improve performance in the problem of trajectory prediction. Additionally, PCGAN and PSGAN achieve comparable or improved performance in 17 out of 30 metrics compared to prior methods suggesting that our probabilistic GAN approach can improve trajectory prediction performance in certain crowd interactions.
  Even when used without vehicle feature input we can see that the inclusion of the GVAT for social pooling in PCGAN improves the performance in the majority of tests compared to PSGAN. However, on both ETH-Hotel and UCY-Univ we find that PSGAN outperforms PCGAN. On these two datasets, CVM also performs well, suggesting that there may be fewer pedestrian interactions involved allowing more linear models to achieve improved results. Sch{\"o}ller et al. \cite{Scholler2019} demonstrated the effectiveness of CVM, and we find that this result still holds even when limited to prediction periods of 8 and 12 timesteps.
SGAN \cite{gupta2018social} performs poorly for all tested datasets when limited to using the average error over multiple samples, as opposed to using the best sample error compared to the ground truth. This result is similar to that obtained in \cite{zhang2019sr},  where SGAN was not found to perform well when limited to a single sample.  Whilst this may be a result of SGAN sampling between multiple future modal paths, our method PSGAN, which extends SGAN for direct probabilistic output, demonstrates that by being able to estimate the likelihood of each modal path, we can greatly decrease the error of the adversarially trained method obtained for all metrics.
Unlike both SGAN and our methods, State Refinement LSTM (SRLSTM) \cite{zhang2019sr} pools across the pedestrian hidden states found from the most recent observation. Whilst only being comparable for predictions of 12 timesteps, this method performs well for all datasets, confirming the importance of using the most recently available information for predictions.

\textbf{Experiment 2:} \textit{Table.\ref{tab:2}} outlines the performance of all compared methods on the VCI and USyd datasets, both of which contain pedestrian-vehicle interactions. These results again highlight how using a probabilistic output during adversarial training can improve prediction results, with both of our methods, PSGAN and PCGAN, improving upon SGAN. Importantly, we can see that by including the vehicle feature input in GVAT pooling we can achieve significant improvements on the VCI dataset, with PCGAN significantly outperforming PSGAN, and outperforming or equalling SRLSTM on all metrics. CVM and PSGAN score well for the MHD metric, suggesting that these methods are likely incorrectly predicting the speed profile of pedestrian trajectories, but correctly predicting the direction.

\subsection{Qualitative Evaluation}
\textbf{Experiment 1:}  
\textit{Fig.\ref{qual_nonvehicle}} demonstrates realistic behaviours between pedestrians, producing results that reflect the actual probabilistic and multimodal nature of crowd interactions. (a) and (b) both reflect the ambiguity expected during an interaction between two pedestrians. The two possible trajectories that can be taken to avoid an oncoming pedestrian are clearly displayed in the modal paths of example (a), where one branch of the modal path tree matches the actual trajectory taken. This situation is again seen in \textit{Fig.\ref{qual_comp}} (top), where PSGAN (dark green) and PCGAN (light green) are able to accurately predict the turning of oncoming pedestrians with branching modal paths, whilst both SRLSTM (pink) and CVM (yellow) do not account for this ambiguity. Likewise, example (b) reflects the possibility that the two pedestrians might continue turning together, or instead travel forwards beside each other. Additional examples extend these ideas to more crowded scenes, with multiple pedestrians displaying the similar multimodal and uncertain interactions.
In \textit{Fig.\ref{qual_nonvehicle}} (b), whilst there exists clear dependency between the two predicted forking modal path trees, our model does not currently have the ability to determine this relationship, and so cannot predict which branch an agent will take even with knowledge of the true path of a neighbouring agent.

\phantom{xx}\textbf{Experiment 2:}
The extension of our approach to include a vehicle allows the modelling of interactions in shared pedestrian-vehicle environments, predicting crowd response in the presence of a vehicle as shown in \textit{Fig.\ref{qual_vehicle}}. Experiment 2 uses a shorter timestep, of only 0.1 second on the USyd, and 0.083 seconds for VCI. As is expected over this shorter time we don't see as significant interactions, reflected in near-linear ground truth in both \textit{Fig.\ref{qual_vehicle}} and \textit{Fig.\ref{qual_comp}} (bottom). However, we can still see clear multimodal predictions in certain interactions, including when the vehicle is approaching pedestrians from behind as in \textit{Fig.\ref{qual_comp}} (a), where the closest pedestrian responds by beginning to move to the side. This interaction is reflected in the predicted modal paths, although the sideways direction is predicted in the wrong direction.  In \textit{Fig.\ref{qual_comp}} (bottom) we also see how only PCGAN accounts for the vehicle's influence on the pedestrians, correctly predicting the possibility that the pedestrians will return to their original motion once the vehicle has passed.

\section{Conclusion}
Our work shows how a direct multimodal probabilistic output can be generated in an adversarial network for pedestrian trajectory prediction, outperforming existing methods including sampling-based approaches. We additionally show how the presence of an autonomous vehicle can be considered through the introduction of a novel GVAT pooling mechanism. By comparing our work to \cite{gupta2018social}, a non-probabilistic GAN used for trajectory prediction, we have shown that our probabilistic approach clearly benefits adversarial training for this problem. 
Our work focuses on how a single vehicle can operate away from the lane-based structure of a road, examining crowd interactions to enable safer decisions, however could in future be extended for use with multiple vehicles through inclusion of all vehicles as nodes, removing the \textit{z} term from \textit{Eq. \ref{eq:nodeinputs}} and replacing $W_r$ and $W_u$ with a different set of weights for each agent type pair to learn relationship dynamics.

{\small
\bibliographystyle{unsrt}
\bibliography{egbib}
}

\end{document}